# Deep Learning based Automatic Detection of Dicentric Chromosome

Angad Singh Wadhwa* Nikhil Tyagi and Pinaki Roy Chowdhury

**ABSTRACT:**
Automatic detection of dicentric chromosomes is an essential step to estimate radiation exposure and development of end to end emergency bio dosimetry systems. During accidents, a large amount of data is required to be processed for extensive testing to formulate a medical treatment plan for the masses, which requires this process to be automated. Current approaches require human adjustments according to the data and therefore need a human expert to calibrate the system. This paper proposes a completely data driven framework which requires minimum intervention of field experts and can be deployed in emergency cases with relative ease. Our approach involves YOLOv4 to detect the chromosomes and remove the debris in each image, followed by a classifier that differentiates between an analysable chromosome and a non-analysable one. Images are extracted from YOLOv4 based on the protocols described by WHO-BIODOSNET. The analysable chromosome is classified as Monocentric or Dicentric and an image is accepted for consideration of dose estimation based on the analysable chromosome count. We report an accuracy in dicentric identification of 94.33% on a 1:1 split of Dicentric and Monocentric Chromosomes.

1. **INTRODUCTION**

Cytogenetic dosimetry (Organization and others, 2011) involves the identification of Dicentric Chromosomes (DCs) for radiation dose estimation on the basis of the frequency of occurrence of the DCs. This information is vital to save the lives of people after a radiation accident. A plan of treatment is formed after the dose of radiation is estimated for each individual. The human chromosome undergoes mutation after being exposed to radiation which involves breaking and subsequent linking of two broken chromosomes. This linkage leads to the formation of a DC from two monocentric (MC) chromosomes. A DC has two centromeres whereas a MC has only one centromere (Fig 1). A human expert is regarded as the gold standard to assess the number DCs and estimate the radiation dose. The scoring process is labour and time intensive. Our aim is to improve the efficiency of the system for mass casualty events by automating the scoring process. Modern advances in Computer Vision have led to development of systems which can score the images in a short period of time (Galloway et al., 2020; Shen et al., 2019). These systems are dependent on certain threshold values according to the quality of images in a particular batch, thus dependent on certain user given empirical values which are subjective in nature. If the image is blur or noisy in a particular batch, or if the size and proximity among chromosomes vary, then the thresholds will have to be tweaked accordingly. This is the cause of subjectivity and therefore it becomes expert dependent. Our work is independent of such subjective values given by the domain expert which can depend on variety of conditions. Due to the decrease in performance, the idea is to remove such subjectivity by using contextual information which in turn is expected to increase the efficiency. That motivates us to develop an efficient and automated model to achieve this goal of higher accuracy consistently.

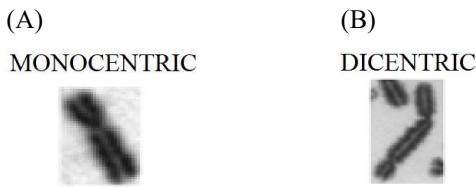

Fig 1: (A) Monocentric Chromosome (MC) (B) Dicentric Chromosome (DC)

Feature extraction based on morphological operators combined with machine learning techniques have been implemented earlier for analysis of DCs but suffer from a trade-off between false positive rate and number of images that can be processed (de Faria et al., 2005; Liu et al., 2017). Previous works have also incorporated certain measures to assess the quality of images like custom filters but still depend on certain threshold values (Liu et al., 2017). A distinguishing factor between MC and DC are the number of centromeres and shape of the chromosome. The pipeline of the previous works involves filtering out bad quality images (Liu et al., 2017) (false positives), chromosome segmentation (de Faria et al., 2005; Shen et al., 2019), centreline extraction (Shen et al., 2019), centromere detection (Subasinghe et al., 2016), edge detection (Yan and Shen, 2008), thresholding algorithms (Karvelis et al., 2008, 2006; Manohar and Gawande, 2017) and geometric algorithms (Minaee et al., 2014). Chromosome extraction is a major issue faced in the previous works due to objects like nuclei, smudges, debris and overlapping chromosomes. Recent developments in deep learning techniques have seen significant progress in segmentation, extraction and classification in different kinds of medical imaging problems (Banerjee et al., 2020; Ronneberger et al., 2015). A deep-learning system can extract features and learn decision thresholds from data without any intervention from domain experts, however, the traditional and conventional systems happen to be time-consuming and costly given the esoteric nature of medical images. Human crafted features and thresholds often fail to generalize well on medical images given the scarcity and variability of medical data.

We propose InceptionResnetv2 based deep learning method for classification of DCs to solve the problems mentioned above with high accuracy as this will facilitate the formulation of medical plans for diseased people objectively, consistently, and in quick time which is repeatable every time with similar efficiency and efficacy. We present comparative studies with other models and show the comparison in terms of efficiency/efficacy and the computation time. We discuss the reasons for the selection of our specific methods and then the pipeline involved in our system, after explaining the shortcomings of the previous studies and certain experiments of our own, which led to the formation of our proposed approach. The paper is organised as follows: Section 2 presents the related work which has close linkages with the technologies that are proposed in this paper. In Section 3 we discuss the technologies that are used in this paper along with results and discussion in Section 4 & 5. We conclude the paper in Section 6.

2. **Related Work**

    **Image Processing:** Previous works following traditional computer vision and image processing techniques use a particular pipeline to detect DCs. Segmentation is done by a thresholding method like watershed (Manohar and Gawande, 2017), or Otsu (Otsu, 1979). The contours with closed areas are detected by a contour finding method (Suzuki and others, 1985) and in some cases a skeleton (Shen et al., 2011) is drawn by skeleton drawing algorithms. The centromere is located through contour width. The contour width of the chromosome is the distance from the skeleton to the contour. Contour width is less wherever the centromere lies as compared to the contour width of the rest of the chromosome. The nuclei and debris are removed through mathematical filters which consider contour area, ratio of adjacent sides of the bounding rectangle and intensity of pixels (Liu et al., 2017). The noise can be removed using Fourier transform (Bracewell and Bracewell, 1986) by eliminating the high frequency components or other such methods. Despite selecting a good quality image, these methods rely on empirical values which are user dependent. In (Shen et al., 2019) it can be observed that the true positive rate (76.6%) and positive predictive value (76.6%) are low. We suspect these low values are due to high number of false positives and false negatives. This can be quite harmful when formulating a plan of medical treatment on the basis of these results. These drawbacks are the main reasons for our deep learning inspired pipeline which is independent of such decision thresholds.

    **Deep Learning:** Deep learning based approaches have been applied to differentiate between analysable and non-analysable quality images of chromosomes for evaluation (Jang and others, 2020). Denoising of chromosomes has been experimented by U-Net (Ronneberger et al., 2015) based models for removal of nuclei, debris, salt and pepper based noise (Bai et al., 2020). Detection and extraction of chromosomes can be done through YOLO and U-Net (Bai et al., 2020) based models respectively but have in turn

hampered image information while extraction (Fig 2). It can be observed visually from Fig 2 which is obtained after extraction, there is loss in overall structure of the chromosome which can lead to incorrect classification. The results did not change even after fine tuning the model. Deep learning has been applied for the purpose of karyotyping (Hu et al., 2019) with an accuracy of 93.79% but detection of DCs is still a persistent problem since it deals with the minute differences in structure which has been challenging due to scarcity of labelled data and variation of data among labs and batches.

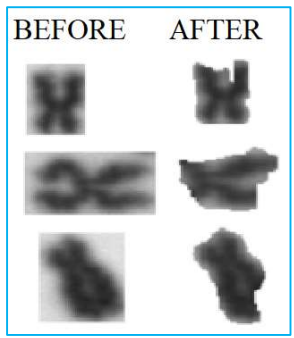

Fig 2: Result of U-NET. The images before and after obtained from the segmentation model show the loss in structural information after extraction and denoising which leads to incorrect chromosome classification

We performed several experiments to detect DCs while keeping the computational complexity, time and accuracy in mind.

3. **Methods**

3.1. **Techniques:**

3.1.1. Principal Component Analysis (PCA):

PCA is linear algebra-based algorithm which is primarily used for dimensionality reduction of tabular data where columns represent different features and rows store data samples. PCA must fit on a training set, before being used to reduce any data. Therefore, training set determines how effective reduced data is for classification. Images can be converted to tabular format by flattening the image into an array of pixels and stacking multiple images vertically to create a dataset. PCA used in such manner can be treated as image feature extractor. PCA can also reconstruct original image (data) from the reduced form. Although such reconstruction is not perfect as the reduction step destroys data to perform dimensionality reduction and reconstruction is dependent on training set. Accuracy of reconstruction can be increased at the expense of increasing dimensionality of reduced data. We leverage the inability of PCA to reconstruct exact replica of input data to create a purely PCA based classifier. We will describe the classifier in an object-oriented paradigm where the classifier has '$n$' number of PCA objects each belonging to the '$n$' number of classes in the data. Each PCA object has three methods which are fit, reduce and reconstruct. Training data is separated into constituent classes and each PCA object is fit on a single class using method 'fit'. For testing we create '$n$' copies of the image and reduce the copies using PCA objects. Each reduced copy is then reconstructed using the same PCA object that reduced it. This creates '$n$' reconstructed images each belonging to their respective classes. Each reconstruction is then compared to the original image using Mean Squared Error (MSE), image belongs to the class whose reconstruction gives the least MSE.

3.1.2. Brain Inspired Facial Recognition (BIFR) based model:

The problem of facial recognition depends on the minute differences that the model can distinguish between faces which will lead to classification. Faces can be quite similar to one another and differentiating between them is a complex task on its own, which is quite similar to the task of dicentric chromosome identification. We utilised the unique property of the model to be trained on less images and to identify these minute differences. LBP (Ojala et al., 1996) and HOG (Dalal and Triggs, 2005) are feature extraction algorithms that work as visual descriptors and are used for extracting features from an image. LBP is a texture operator that detects minute changes in local spatial patterns and is invariant to monotonic gray-scale changes. HOG detects the intensity gradients or edge directions in an image and extracts these spatial features in the form of a vector. PCA is used as a feature extractor and a dimensionality reduction operator for LBP and HOG in the pipeline of BIFR based chromosome classification. These features are given to a sparsely connected multi-layered perceptron (MLP) and fused at the decision level to finally classify an image (Chowdhury et al., 2021).

### 3.1.3. CNN and YOLOv4:

Convolutional neural networks (CNN) are state of the art image processing algorithms which can be used for classification, segmentation and detection. CNNs outperform traditional computer vision algorithms in image analysis with enough data. YOLO (Redmon et al., 2015) or You Only Look Once is an object detection algorithm which is a single stage detector that creates bounding boxes (Fig 3 (A)) using intersection over union and non-maximum suppression. We use YOLOv4 (Wang et al., 2020) to detect the chromosomes in our pipelines. InceptionResnetv2 is a CNN used for classification of analysable, non-analysable and MC and DC classification.

### 3.1.4. Data Augmentation:

Data augmentation techniques were applied to increase the training data. Initially we label the images using open-source Computer Vision Annotation Tool (CVAT) by drawing bounding boxes on chromosomes for training YOLOv4. We create a pool of individual chromosomes and debris to choose from and spread those individual images on white background which can be seen in Fig 3 (B). Otsu thresholding was applied on good quality images with spaced out chromosomes to extract individual chromosomes. These masks formed were then extracted using contour finding algorithm (Suzuki and others, 1985). The extracted chromosomes and debris are randomly used to form an artificial image. These artificial images (with labels) were used to increase the training data for YOLOv4. The data augmentation technique that we use in this paper is a combination of random flip and gamma variations. We experiment with other augmentation methods like affine transformations, elastic transformations and piecewise affine transformation. Affine transformations include translation and scaling on x axis and y axis, rotation and shear.

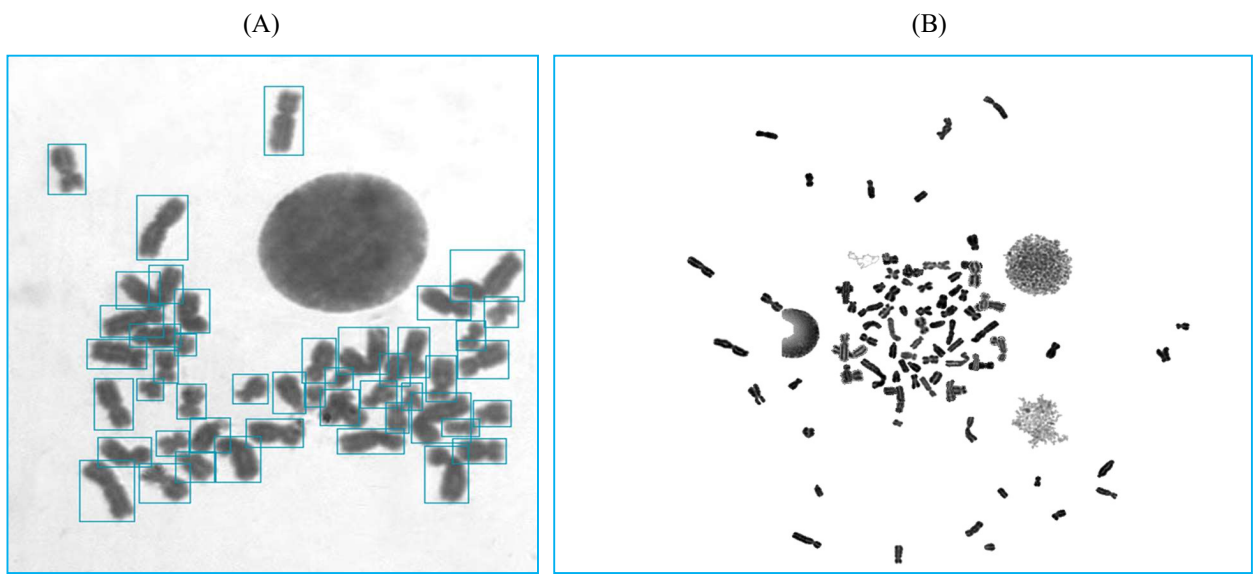

Fig 3 : (A) Result of YOLOv4 (B) Artificially created images

## 3.2. Pipelines:

### 3.2.1. Blending Potential Ratio (BPR) based method:

Initially, the image is denoised using Fourier transform (see Fig 4 (B)). The high frequency component of the image is removed in the Fourier domain and then inverse Fourier transform is applied to convert the image back to the image domain. A binary mask of the image is formed using Otsu threshold (see Fig 4 (C) and Fig 5 (C)). Active Contours Without Edges-ACWE (Chan and Vese, 2001) has also been experimentally tested for thresholding but there is no major difference visually (see Fig 5). The filters (Fig 4 (E)) are used to remove debris which are left after denoising that are ratio of - 1) area and median area, 2) mean width and median of mean width, 3) median width and median of all the median widths, 4) maximum width and median of maximum width and 5) minimum length of bounding rectangle and maximum length. For further detail, the readers are referred to the paper (Liu et al., 2017). These filters are applied on the contours (Fig 4(D)) of the binary mask of the image. The workflow of this pipeline can be seen in Fig 6 (E).

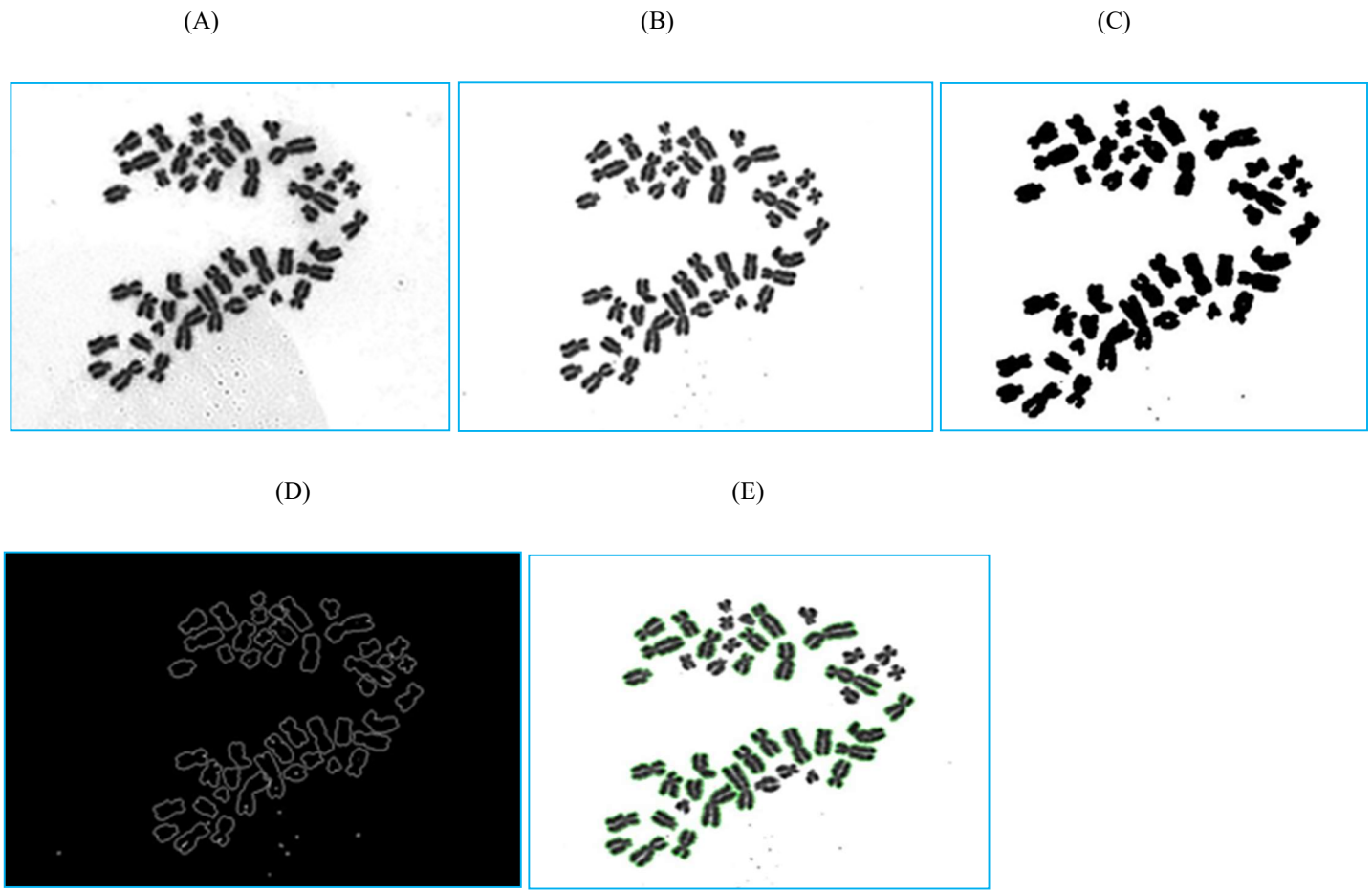

Fig 4: (A) Original Image : (B) Image after Fourier Transform and denoising : (C) Thresholded Image : (D) All Contours (E) Contours after applying filters

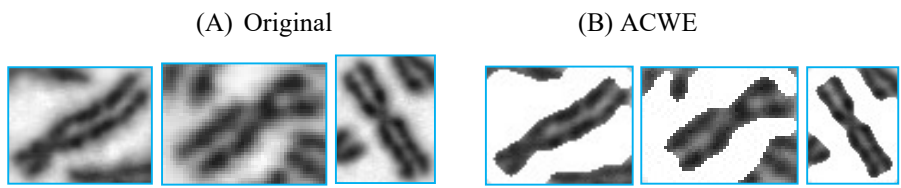

(A) Original  (B) ACWE

(C) OTSU

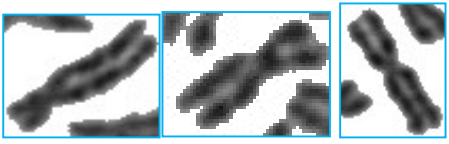

Fig 5: (A) Original image : (B) Images after applying Active Contours Without Edges – ACWE method : (C) Image after applying OTSU thresholding

The Blending Potential Ratio (BPR) method (Shen et al., 2011) is used to find the centromeres. A visual example of the method is given in Fig 6 (A), (B), (C), (D). BPR is used in the process of skeletonization (Fig 6 (B). Modifying the algorithm, the DCs can be calculated. The point which forms the skeleton is called the skeleton point. The following equation calculates the metric on which we decide if a point is a skeletal point or not. Centromere candidates are chosen from skeletal points.

$$\varepsilon(p_n, q1, q2) = \tan\frac{\theta}{2} * sqrt(\frac{len(q1,q2)^2}{dist(q1,q2)^2} - 1), \quad (1)$$

where '$p_n$' is the point that is being investigated, '$q1$' is the boundary point nearest to '$p_n$' (called the ruling point of "$p_n$") and '$q2$' is the ruling point of one of $p_n$'s eight connected neighbours (Fig 6 (D)).

'$\theta$' is the angle between $q1$, $p_n$ and $q2$. The minimum length of the boundary segments created by $q1$ and $q2$ is given by $len(q1, q2)$. The Euclidean distance between $q1$ and $q2$ is given by $dist(q1, q2)$. A point '$p_n$' is a skeletal point if $\varepsilon(p_n, q1, q2) > t$ where $t$ is a threshold usually between (0.5-1.5) and $q1$, $q2$ are with respect to $p_n$.

For each contour find two skeletal points $p_1$ and $p_2$ such that they have minimum $(dist(p_n, q1) + dist(p_n, q2))$ of all skeletal points and $dist(p_n, q1) + dist(p_n, q2) - dist(q1, q2) < 0.1$, where, $p_n$ is replaced by $p_1$ and $p_2$ such that $p_1$ has minimum $(dist(p_n, q1) + dist(p_n, q2))$ out of both. $p1$ and $p2$ are possible centromere candidates. If $dist(p_1, p_2) > tl$, then $p_2$ is accepted as a candidate else rejected. '$tl$' is calculated empirically by taking the average length of chromosomes in a cell (approximately 46) into consideration, for example if the average length of the chromosomes is 50 pixels, then the distance can be 5 pixels (approximately 10%)

If the ratio of $\frac{dist(p_2,q1)+dist(p_2,q2)}{dist(p_1,q1)+dist(p_1,q2)} > 1.05$ only $p_1$ is selected as centromere else both are selected as centromere. The threshold 1.05 is obtained empirically. Chromosomes with 2 centromeres are classified as DC and those with one centromere are classified as MC.

(A)　　　　　　　　　　　　　(B)　　　　　　　　　　　　　(C)

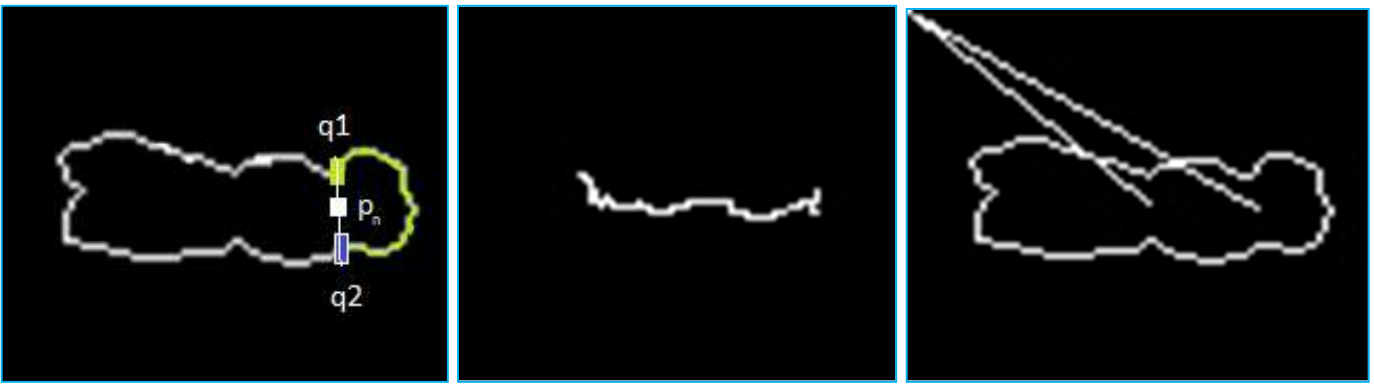

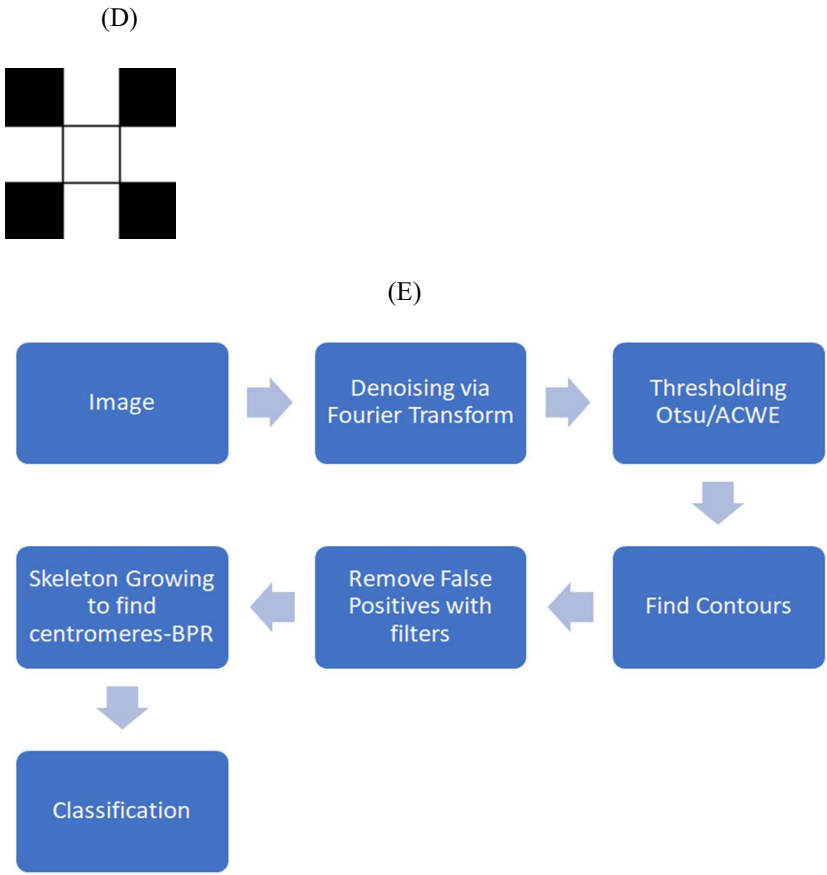

Fig 6: (A) Boundary of chromosome, where the white dot inside is the point $p_n$, blue point on the bottom is $q2$, green point on the top is $q1$ and boundary marked in colour green is *len(q1,q2)* : (B) Skeleton of chromosome : (C) Detected Centromeres in a chromosome : (D) A pixel and it's eight connected neighbours : (E) Workflow for BPR Classifier. The input image is denoised via Fourier transform and image thresholding is applied via OTSU/ACWE (active contours without edges) to find the contours. Mathematical filters are applied to remove the false positives and a skeleton growing algorithm-Blending Potential Ratio (BPR) is applied to find the skeleton. The centromeres are found from the minimum width from the skeleton to the chromosome contour as a chromosome is narrow near the centromere.

3.2.2.    Brain Inspired Face Recognition (BIFR) model for Chromosomes : The input image is processed through YOLOv4 (object detection model) (Wang et al., 2020) which is used to extract the individual chromosomes. A detailed explanation of training and data preparation for YOLOv4 is presented above in section 3.1.3. This pipeline uses BIFR (Chowdhury et al., 2021) model as the primary classifier for image rejection followed by MC and DC classification for chromosomes. These features are given as input to a (MLP) which then classifies the image as analysable or non-analysable initially and then as MC or DC. The workflow of the BIFR pipeline can be seen in Fig 7.

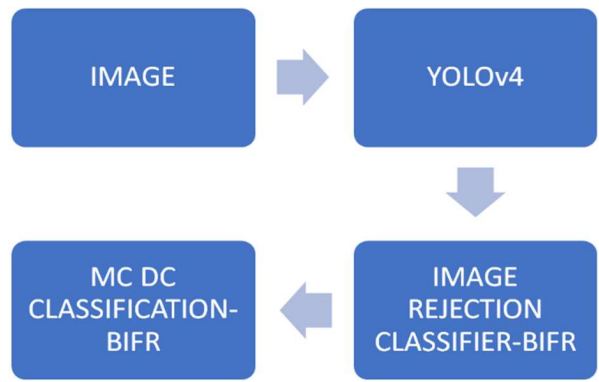

Fig 7: Workflow for DC detection using the BIFR Classifier. An image is taken as input which is then processed via YOLOv4 to detect and extract the chromosomes. The chromosomes are then passed through BIFR based model for image rejection which accepts or rejects an image. The accepted chromosomes are then classified by another BIFR based model for DC detection and an image is accepted based on the final chromosome count.

3.2.3. PCA Classifier:

YOLOv4 is used to extract the individual chromosomes. We then split training data into its respective constituent classes which are analysable & non-analysable in the case of image rejection and monocentric & dicentric in the case of dicentric identification. Two PCA class instances are then initialized each for the two classes i.e., analysable and non-analysable. PCA (Wold et al., 1987) objects are fitted on their respective class training datasets. Entire testing data is transformed and inverse-transformed by both PCA objects separately. L2 norm distance from both instances of reconstruction to real test data is calculated. Image is assigned to the class which has the least distance. This process is repeated once again for classes MC and DC (Fig 8).

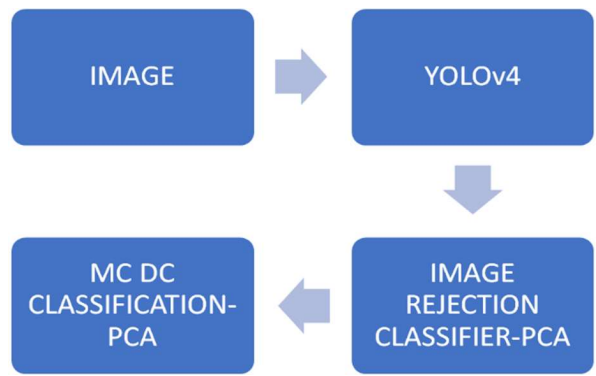

Fig 8: Workflow for PCA based Classifier for DC detection. An image is taken as input which is then processed via YOLOv4 to detect and extract the chromosomes. The chromosomes are then passed through PCA based classifier which accepts and rejects an image. The accepted chromosomes are then classified by another PCA based classifier for dicentric detection and an image is accepted or rejected based on the final chromosome count.

3.2.4. Deep Learning Framework:

Initially we label the images using open-source Computer Vision Annotation Tool (CVAT) by drawing bounding boxes on chromosomes for training YOLOv4. We apply transfer learning and use pre trained weights of MS COCO dataset (Lin et al., 2015). To increase the number of training images for YOLOv4 we use the method described in section 3.1.4. YOLOv4 is used to extract the individual chromosome images which are further processed in the pipeline. A workflow of this pipeline can be seen in Fig 9. Training images for the image rejection classifier were segregated manually based on the protocols described by WHO-BIODOSENET. An analysable image must have a well spread metaphase, sharp and clear morphological chromosome structure, separate chromatids, non-overlapping chromosomes and no residual cytoplasm. A non-analysable image is blur, morphologically unclear, has overspread cells, overlapping cells, cytoplasmic cells, unclear centromeres, condensed chromosomes, pulled apart chromatids, unclear second centromere, twisted chromatids, multiple chromosomes in a straight lines and swollen chromosomes. These images were manually segregated. MC and DC images were further segregated for the dicentric classifier to classify the images. We extract small patches of the entire slide containing a cluster of chromosomes from the same cell typically of size 1024×1280 pixels at 1000× magnification. Each image is then padded to have a common size of 192×192 pixels. Images are then processed by an Image Rejection Classifier (IRC). We created a dataset of 7000 bad images and 15000 good images. The non-analysable images are augmented using augmentation method described in section 3.1.4 to match the number of analysable images. The Training-Testing split was kept at 9:1 for IRC. We train an Inception-Resnet-V2 (Szegedy et al., 2016) pretrained on ImageNet

(Russakovsky et al., 2015) dataset with a learning rate of 1e-5. An example of a non-analysable and analysable chromosome can be seen in Fig 10 (A), (B).

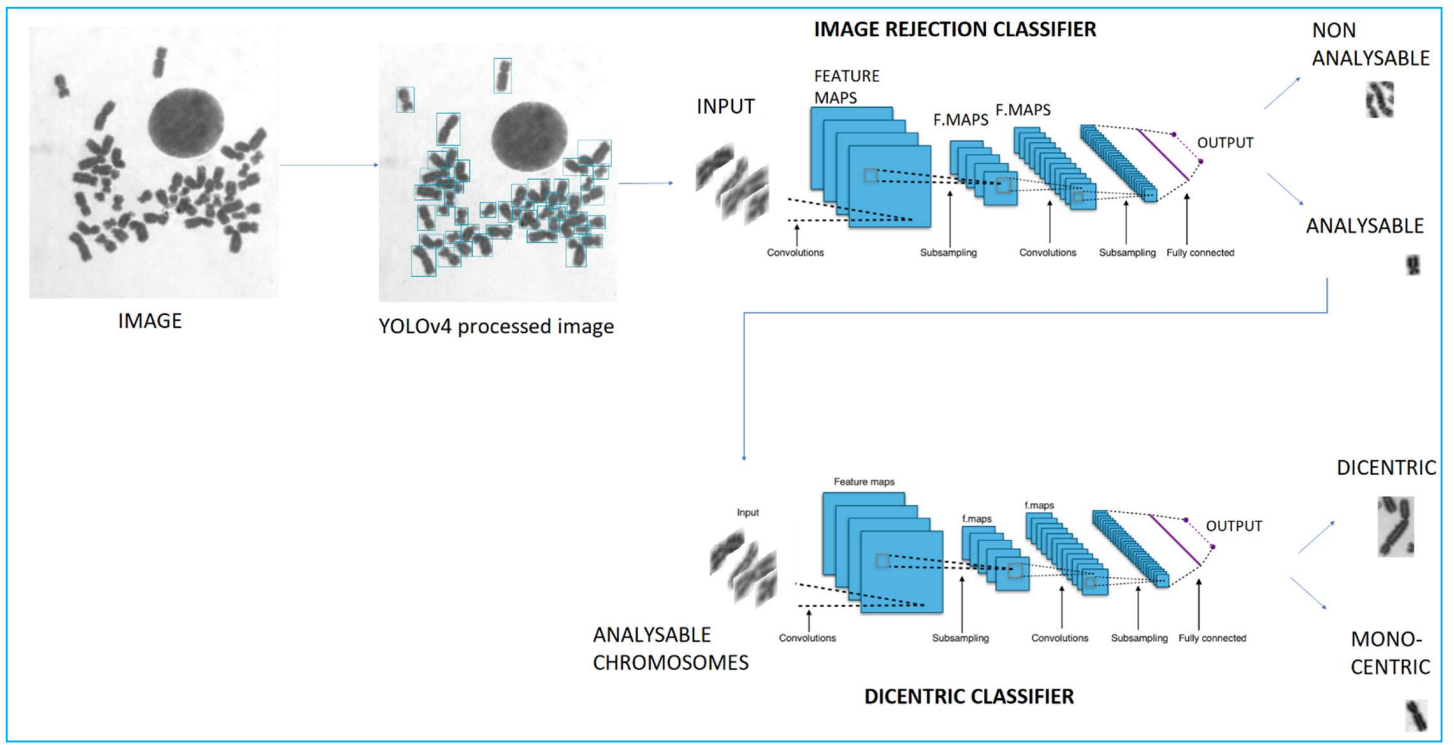

Fig 9: Workflow for InceptionResnetv2 based Classifier for DC detection. An image is taken as input which is then processed via YOLOv4 to detect and extract the chromosomes. The chromosomes are then passed through InceptionResnetv2 based classifier which accepts and rejects an image. The accepted chromosomes are then classified by another InceptionResnetv2 based classifier for dicentric detection and an image is accepted or rejected based on the final chromosome count.

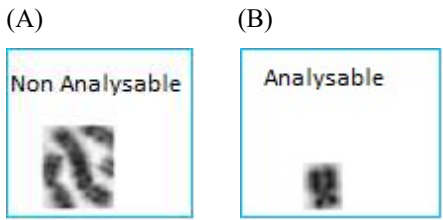

Fig 10: (A) Non-Analysable chromosome (B) Analysable chromosome

For DC classification, we separated 8000 MC and 8000 augmented dicentrics into two classes. The training-testing split for DC classification was kept at 1:1. We train an Inception-Resnet-V2 pretrained on ImageNet dataset (Deng et al., 2009) for 300 Epochs at a learning rate of 1e-4 using AdamW (Kingma and Ba, 2014) as optimizer. Images rejected by IRC are discarded, and the rest are analysed by DC classifier for classification. Any patch which doesn't have a prespecified number of chromosomes is rejected, this process is done before and after IRC.

4. Results

All experiments were performed on 4.5-5gy images of WHO-BIODOSENET. The confusion matrix is calculated which includes the values of true positives (TP), true negatives (TN), false positives (FP) and false negatives (FN), where positive is DC and negative is MC. We report the results on metrics like accuracy, precision, recall, specificity, and Mathews corelation coefficient (MCC) where,

Accuracy is calculated by: $\frac{TP+TN}{TP+FN+TN+F}$ (2)

Precision is calculated by: $\frac{TP}{TP+}$ (3)

Recall is calculated by: $\frac{TP}{TP+FN}$ (4)

Specificity is calculated by: $\frac{TN}{TN+FP}$ (5)

Mathews Corelation Coefficient is calculated by: $\frac{(TP \times FN)-(FP \times FN)}{\sqrt{((TP+FP)(TP+FN)(TN+FP)(TN+FN)}}$ (6)

We report 94.33% accuracy, 98.54% precision, 90% recall, 98.66% specificity and 0.89 mcc of InceptionResnetv2 on a sample of 3000 test images (see Fig 11). Our model outperforms other methods and can be observed from Fig 11 and Fig 12. InceptionResnetv2 performs better and is also computationally more intensive than most of the methods. One can argue that in the BIFR based model the trade-off between accuracy and time is less as compared to InceptionResnetv2. The reason for choosing InceptionResnetv2 based model is of developing a system of high accuracy as this field requires a medical treatment plan to be formulated accordingly.

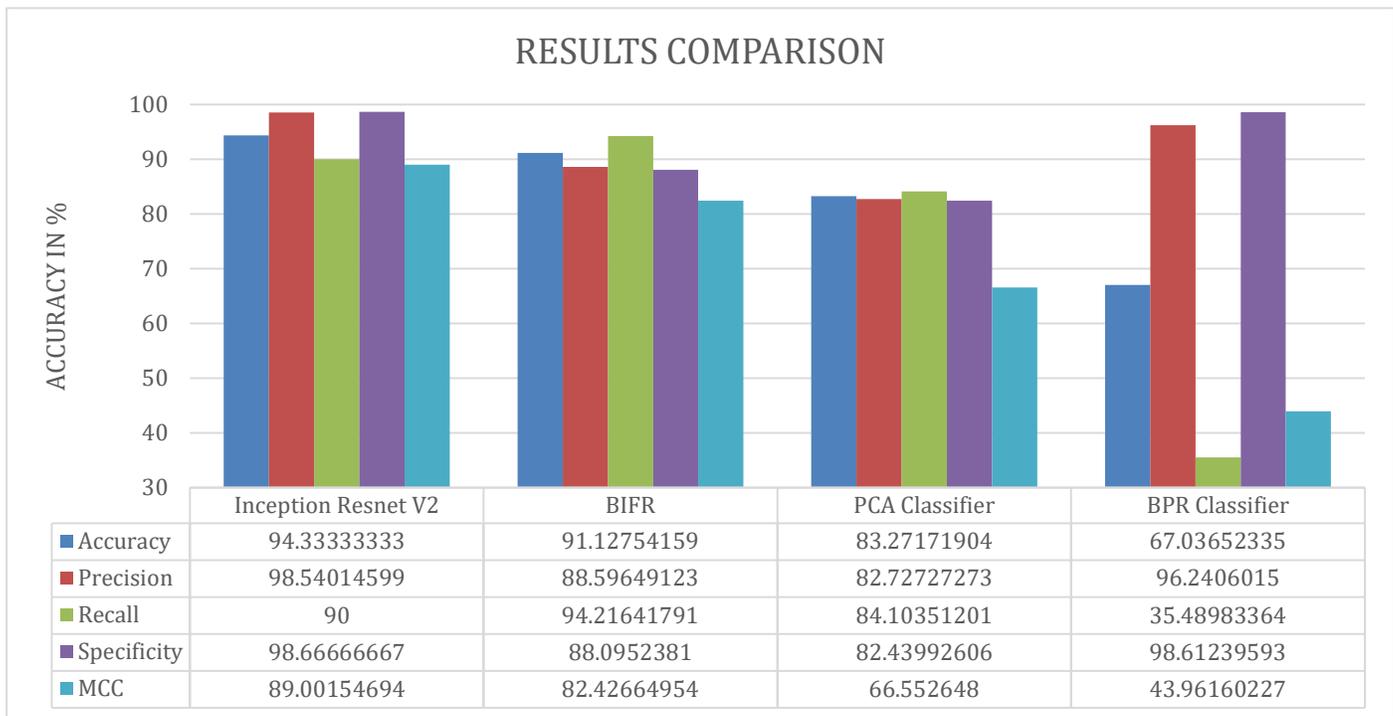

Fig 11: Comparison of proposed pipelines with others in terms of accuracy, precision, recall, specificity and Mathews Correlation Coefficient (MCC).

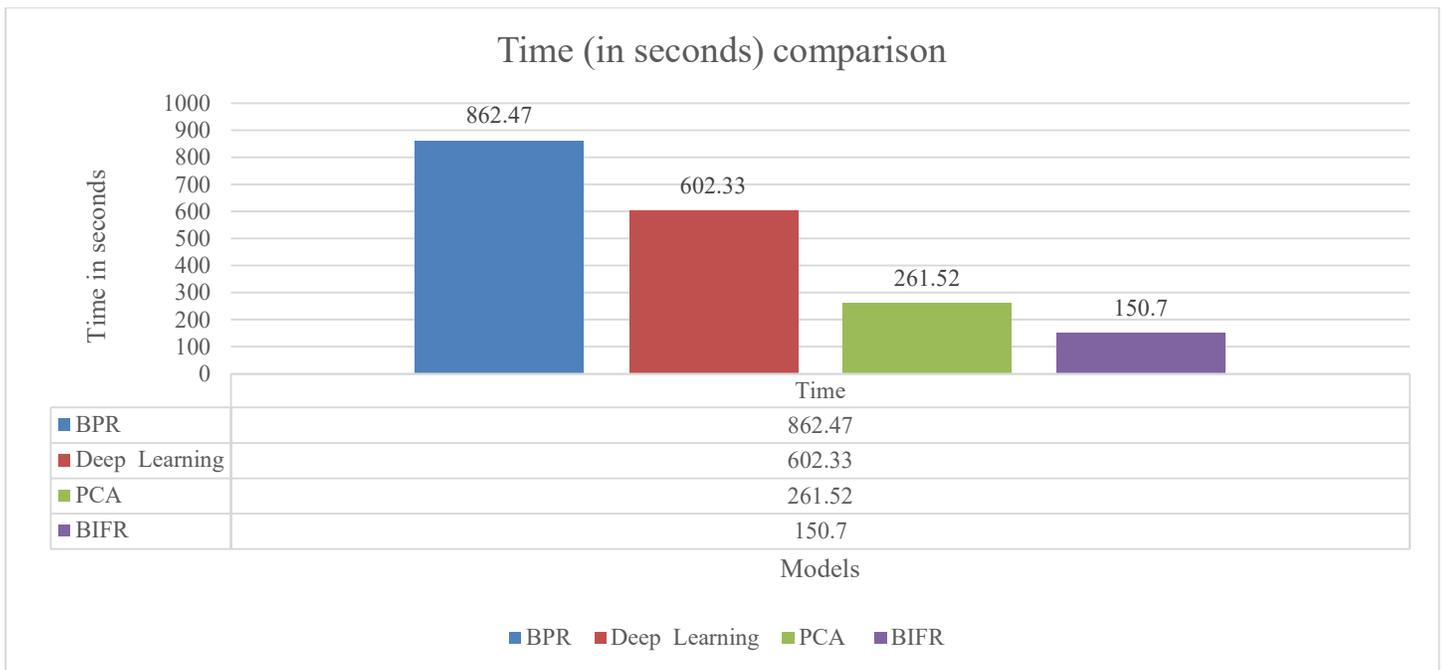

Fig 12: Comparison of proposed pipeline with others in terms of test time (run time) in seconds for 14000 images on CPU.

## 5. Discussion

The BPR based method yield excellent results if the chromosomes are spaced out but yields clusters of chromosomes if the chromosomes are closer together, which is often the case in chromosome patches. We separate clusters by again using Otsu or ACWE on clusters. Clusters left after this step are considered as debris and discarded. The method works for about 80% of the cases but for remaining cases it over separates individual chromosomes, and thus it was not preferred. This image processing-based method is highly dependent on the accurate skeleton drawing and pruning (as explained in section 3.2.1) which in many cases could be incorrect as chromosomes are not always in a straight shape. Even after enhancing the contours using a contour finding technique Sobel (Irwin and others, 1968) the skeleton was not always formed correctly which led to misclassification. Thus, a deep learning-based classifier was implemented to overcome this shortcoming. The data for training this classifier was manually segregated and augmented. One important point that needs to be emphasized here is the cluster separation which was not possible using Otsu thresholding or ACWE due to the minimal difference in space between chromosomes for different images since these methods work on certain thresholds and cannot be approximated for many use cases. YOLOv4 overcame this problem by forming the bounding boxes efficiently.

Since feature extraction played an important role, we experimented with operators like LBP, HOG, ACWE for contours. These features would then be dimensionally reduced with PCA (or any other dimensionality reduction technique) and classified by MLP but the purpose of generalisation could not be satisfied. This experiment yielded the second-best result and can be considered for computationally less intensive environments where extremely high accuracy is not the concern. The results can be improved with more training data. The concept of clustering chromosomes in different clusters based on their sizes and types was experimented via *k-means* clustering (Hartigan and Wong, 1979) but this approach required a threshold to specify the number of clusters to be formed. To overcome this threshold we experimented with *Growing Neural Gas* (Fritzke and others, 1995) which requires no arbitrary value for the creation of clusters. *Growing Neural Gas* and *k-means* were clustering the images on the basis of size and angle of chromosomes. The purpose for clustering was to separate analysable and non-analysable chromosomes but due to the structural similarity in each chromosome, the clusters formed could not separate the chromosomes as expected. This led to the formation of InceptionResnetv2 based deep learning pipeline which required a lot of labelled data but yielded high accuracy which is our prime objective in this case.

We used Python3, keras and Tensorflow for our experiments. All experiments were carried out on our system with Intel i7 10700k, 32 GB RAM

6. **Conclusion**

The task of image recognition that is modelled as a learning classification problem relies on quality and quantity of training data. The scarcity of data for accurate identification and classification of DCs along with the extremely varied and complex structure makes this task hard for even a deep learning network. The final DC counts can be used for dose estimation. This paper is the first attempt to classify DCs based on deep learning methods. To completely automate the task, we need to look forward in the direction of using deep learning networks for this task. We hope that in future the use of disentangled variational autoencoders for classification with sufficient labelled data can improve the pipeline's performance.

**Acknowledgement:**

We thank Dr David Endesfelder at Federal Office for Radiation Protection, BfS Munich for generously helping us by providing the data required for experimentation.

**Funding:**

This research did not receive any specific grant from funding agencies in the public, commercial, or not-for-profit sectors.

**References**

Bai, H., Zhang, T., Lu, C., Chen, W., Xu, F., Han, Z.-B., 2020. Chromosome Extraction Based on U-Net and YOLOv3. IEEE Access 8, 178563–178569.

Banerjee, S., Magee, L., Wang, D., Li, X., Huo, B.-X., Jayakumar, J., Matho, K., Lin, M.-K., Ram, K., Sivaprakasam, M., others, 2020. Semantic segmentation of microscopic neuroanatomical data by combining topological priors with encoder--decoder deep networks. Nat. Mach. Intell. 2, 585–594.

Bracewell, Ronald Newbold, Bracewell, Ronald N, 1986. The Fourier transform and its applications. McGraw-Hill New York.

Chan, T.F., Vese, L.A., 2001. Active contours without edges. IEEE Trans. image Process. 10, 266–277.

Chowdhury, P.R., Wadhwa, A., Tyagi, N., 2021. Brain Inspired Face Recognition System. CoRR abs/2105.0.

Dalal, N., Triggs, B., 2005. Histograms of oriented gradients for human detection. 2005 IEEE Comput. Soc. Conf. Comput. Vis. Pattern Recognit. 1, 886–893 vol. 1.

de Faria, E.R., Guliato, D., de Sousa Santos, J.C., 2005. Segmentation and centromere locating methods applied to fish chromosomes images, in: Brazilian Symposium on Bioinformatics. pp. 181–189.

Deng, J., Dong, W., Socher, R., Li, L.-J., Li, K., Fei-Fei, L., 2009. Imagenet: A large-scale hierarchical image database, in: 2009 IEEE Conference on Computer Vision and Pattern Recognition. pp. 248–255.

Fritzke, B., others, 1995. A growing neural gas network learns topologies. Adv. Neural Inf. Process. Syst. 7, 625–632.

Galloway, S., Coetzer, J., Muller, N., 2020. Image processing-based identification of dicentric chromosomes in slide images, in: 2020 International SAUPEC/RobMech/PRASA Conference. pp. 1–6.

Hartigan, J.A., Wong, M.A., 1979. Algorithm AS 136: A k-means clustering algorithm. J. R. Stat. Soc. Ser. c (applied Stat. 28, 100–108.

Hu, X., Yi, W., Jiang, L., Wu, S., Zhang, Y., Du, J., Ma, T., Wang, T., Wu, X., 2019. Classification of metaphase chromosomes using deep convolutional neural network. J. Comput. Biol. 26, 473–484.

Irwin, F.G., others, 1968. An isotropic 3x3 image gradient operator. Present. Stanford AI Proj. 2014.

Jang, S.S., others, 2020. Automatic Discriminator of Abnormal Chromosomes Using Deep Learning Algorithms.

Karvelis, P.S., Fotiadis, D.I., Georgiou, I., Syrrou, M., 2006. A watershed based segmentation method for multispectral chromosome images classification, in: 2006 International Conference of the IEEE Engineering in Medicine and Biology Society. pp. 3009–3012.

Karvelis, P.S., Tzallas, A.T., Fotiadis, D.I., Georgiou, I., 2008. A multichannel watershed-based segmentation method for multispectral chromosome classification. IEEE Trans. Med. Imaging 27, 697–708.

Kingma, D.P., Ba, J., 2014. Adam: A method for stochastic optimization. arXiv Prepr. arXiv1412.6980.

Lin, T.-Y., Maire, M., Belongie, S., Bourdev, L., Girshick, R., Hays, J., Perona, P., Ramanan, D., Zitnick, C.L., Dollár, P., 2015. Microsoft COCO: Common Objects in Context.

Liu, J., Li, Y., Wilkins, R., Flegal, F., Knoll, J.H.M., Rogan, P.K., 2017. Accurate cytogenetic biodosimetry through automated dicentric chromosome curation and metaphase cell selection. F1000Research 6.

Manohar, R., Gawande, J., 2017. Watershed and Clustering Based Segmentation of Chromosome Images, in: 2017 IEEE 7th International Advance Computing Conference (IACC). pp. 697–700. https://doi.org/10.1109/IACC.2017.0145

Minaee, S., Fotouhi, M., Khalaj, B.H., 2014. A geometric approach to fully automatic chromosome segmentation, in: 2014 IEEE Signal Processing in Medicine and Biology Symposium (SPMB). pp. 1–6.

Ojala, T., Pietikäinen, M., Harwood, D., 1996. A comparative study of texture measures with classification based on featured distributions. Pattern Recognit. 29, 51–59.

Organization, W.H., others, 2011. Cytogenetic dosimetry: applications in preparedness for and response to radiation emergencies.

Otsu, N., 1979. A threshold selection method from gray-level histograms. IEEE Trans. Syst. Man. Cybern. 9, 62–66.

Redmon, J., Divvala, S.K., Girshick, R.B., Farhadi, A., 2015. You Only Look Once: Unified, Real-Time Object Detection. CoRR abs/1506.0.

Ronneberger, O., Fischer, P., Brox, T., 2015. U-Net: Convolutional Networks for Biomedical Image Segmentation. CoRR abs/1505.0.

Russakovsky, O., Deng, J., Su, H., Krause, J., Satheesh, S., Ma, S., Huang, Z., Karpathy, A., Khosla, A., Bernstein, M., Berg, A.C., Fei-Fei, L., 2015. ImageNet Large Scale Visual Recognition Challenge. Int. J. Comput. Vis. 115, 211–252. https://doi.org/10.1007/s11263-015-0816-y

Shen, W., Bai, X., Hu, R., Wang, H., Latecki, L.J., 2011. Skeleton growing and pruning with bending potential ratio. Pattern Recognit. 44, 196–209.

Shen, X., Qi, Y., Ma, T., Zhou, Z., 2019. A dicentric chromosome identification method based on clustering and watershed algorithm. Sci. Rep. 9, 1–11.

Subasinghe, A., Samarabandu, J., Li, Y., Wilkins, R., Flegal, F., Knoll, J.H.M., Rogan, P.K., 2016. Centromere detection of human metaphase chromosome images using a candidate based method. F1000Research 5, 1565.

Suzuki, S., others, 1985. Topological structural analysis of digitized binary images by border following. Comput. vision, Graph. image Process. 30, 32–46.

Szegedy, C., Ioffe, S., Vanhoucke, V., Alemi, A., 2016. Inception-v4, Inception-ResNet and the Impact of Residual Connections on Learning.

Wang, C.-Y., Bochkovskiy, A., Liao, H.-Y.M., 2020. Scaled-YOLOv4: Scaling Cross Stage Partial Network. CoRR abs/2011.0.

Wold, S., Esbensen, K., Geladi, P., 1987. Principal component analysis. Chemom. Intell. Lab. Syst. 2, 37–52.

Yan, W., Shen, S., 2008. An edge detection method for chromosome images, in: 2008 2nd International Conference on Bioinformatics and Biomedical Engineering. pp. 2390–2392.